\pdfoutput=1

\documentclass[11pt]{article}

\usepackage[final]{acl}

\usepackage{times}
\usepackage{latexsym}
\usepackage{textcomp}
\usepackage[T1]{fontenc}
\usepackage[utf8]{inputenc}
\usepackage{microtype}
\usepackage{inconsolata}
\usepackage{booktabs}
\usepackage{xspace}
\usepackage{amsmath}
\usepackage{amssymb}
\usepackage{mathtools}
\usepackage{tcolorbox}
\usepackage{xspace}
\usepackage{caption}
\usepackage{subcaption}
\usepackage{balance}
\usepackage[hang,flushmargin]{footmisc}
\usepackage{multirow}
\usepackage[title]{appendix}

\newcommand{\ourmodel}{\textsc{Drama}}

\title{\ourmodel{}: Diverse Augmentation from Large Language Models \\ to Smaller Dense Retrievers}

\author{
Xueguang Ma$^{2,}$\thanks{Equal contribution. $^\dagger$Work done while at Meta.}$^{,\dagger}$\quad
Xi Victoria Lin$^{1}$\quad
Barlas Oguz$^{1}$\quad
Jimmy Lin$^{2}$\\
\textbf{Wen-tau Yih}$^{1}$\quad
\textbf{Xilun Chen}$^{1,*}$\\
[1ex]
$^{1}$FAIR at Meta\quad
$^{2}$University of Waterloo\quad \\
[1ex]
\texttt{x93ma@uwaterloo.ca},\quad \texttt{xilun@meta.com}
}

\begin{document}
\maketitle

\begin{abstract}

Large language models (LLMs) have demonstrated strong effectiveness and robustness when fine-tuned as dense retrievers.
However, their large parameter size presents significant computational challenges at inference time.
While smaller retrievers offer better efficiency, they often fail to generalize effectively with limited supervised fine-tuning data.
In this work, we introduce \ourmodel{}, a training framework that leverages LLMs to train smaller generalizable dense retrievers.
In particular, we adopt pruned LLMs as the backbone and train on diverse LLM-augmented data in a single-stage contrastive learning setup.
Experiments show that \ourmodel{} offers better multilingual and long-context capabilities than traditional encoder-based retrievers, and achieves strong performance across multiple tasks and languages.\footnote{Code and checkpoints will be available at \url{https://github.com/facebookresearch/dpr-scale/tree/main/drama}.}


\end{abstract}

\section{Introduction}
Recent advancements in large language models (LLMs) have demonstrated their effectiveness and robustness in text retrieval tasks~\citep{muennighoff2024generative, sun-etal-2023-chatgpt, li2024making, behnamghader2024llmvec, lee2025nvembed}.
Directly fine-tuning advanced billion-parameter LLMs with available annotated data can generalize significantly better than fine-tuning a pre-LLM-era smaller model with only a few hundred million parameters~\citep{ma2024repllama, luo-etal-2024-large}.
However, the large parameter size of LLMs brings non-negligible inference-time compute costs, such as encoding large-scale corpora and increased query latency.
For example,
using Llama3.1$_\text{8B}$~\citep{llama3} as the backbone increases the inference cost nearly 40$\times$ compared to a dense retriever based on BERT.


In this work, we holistically explore how to effectively leverage LLMs to create smaller retrievers, in terms of both \textit{data} and \textit{model backbone}, to develop generalizable yet efficient dense retrievers with fewer than 1B parameters.

Although several works have discussed using LLMs for retrieval data augmentation, such as directly generating training triplet~\citep{wang-etal-2024-improving-text} or 
using LLM to mine positive and negative documents from a real corpus~\citep{lee2024gecko}, the effectiveness of these methods has not been rigorously compared.
We comprehensively study the effectiveness of multiple LLM data augmentation methods with a controlled setup: using the same models and corpora across different data creation methods and only relying on open-sourced models and open-access data.
Specifically, we utilize an LLM retriever based on Llama3.1$_\text{8B}$ and an instruction-tuned LLM based on Llama3.3$_\text{70B}$-Instruct to generate augmentation data.
This includes computationally cheap approaches such as generating cropped sentences as queries and using an LLM retriever to mine positive and negative documents over a corpus, as well as computationally expensive methods that further utilize Instruct-LLMs to generate queries and provide relevance judgment.
We investigate the effectiveness of various combinations of these diverse LLM augmentations, providing high-quality augmented training data for English and multilingual retrieval.

Existing smaller dense retriever models are mostly based on encoder-only architectures 
~\citep{wang-etal-2023-simlm,chen-etal-2024-m3,warner2024modernbert}.
We instead propose to leverage LLMs as the backbone for dense retrievers by pruning the decoder-only LLM into a smaller size to initialize the text encoder.
Specifically, we further prune Llama3.2$_\text{1B}$ (which is pruned from Llama3.1$_\text{8B}$) into 0.1B (on par with BERT-base) and 0.3B (on par with XLM-RoBERTa-Large), while preserving multilingual and long-context capability.
We demonstrate that by simply turning on the bi-directional attention during retriever training, the pruned decoder-only models perform well as retrievers.
This offers a more flexible way to create smaller dense retrievers with arbitrary sizes while still leveraging pretrained LLM weights, making smaller retrievers compatible with current and future LLM advancements.

Combining LLM-based data augmentation and backbones, we introduce a single-stage training framework: \ourmodel{} (smaller \underline{D}ense \underline{R}etriever from diverse LLM \underline{A}ug\underline{M}ent\underline{A}tion).
Our smaller retriever models achieve high performance on BEIR~\citep{thakur2021beir}, MIRACL~\citep{miracl}, and multiple multilingual retrieval tasks on MTEB~\citep{muennighoff2022mteb}, outperforming traditional encoder-based retrievers that rely on multi-stage contrastive learning.
This demonstrates that our training framework produces models that excel across diverse English retrieval tasks and exhibit robust multilingual performance, showing the potential for unified smaller retrievers that perform effectively across tasks and languages.

In summary, our contributions are as follows:
\vspace{-2mm}
\begin{itemize}
    \item We investigate diverse methods for leveraging LLMs to generate data augmentation for training smaller models, analyzing their individual and combined effectiveness.
\vspace{-2mm}
    \item We prune LLMs to derive smaller decoder-only language models as backbones for retrievers, demonstrating their advantages in generalizability, multilingual effectiveness, and long-context extrapolation.
\vspace{-2mm}
    \item Our training framework produces a series of multilingual and generalizable smaller retrievers, highlighting the benefits of aligning smaller retriever training with ongoing advancements in LLMs.
\end{itemize}

\section{Related Work}
\subsection{Robust Dense Retrieval}
Dense Passage Retrieval~\citep{karpukhin-etal-2020-dense} utilizes a pre-trained language model such as BERT~\citep{devlin-etal-2019-bert}, to encode text into dense vectors and conduct passage retrieval as a nearest neighbor search.
This approach has shown strong in-domain effectiveness compared to traditional lexical retrievers such as BM25~\citep{robertson2009bm25}.
However, dense retrievers have been found to struggle with generalization when applied to out-of-domain retrieval tasks~\citep{thakur2021beir}.
To address this issue, various works have aimed to improve the generalization of dense retrievers through continuous pre-training tailored for retrieval tasks.

Works such as Condenser~\citep{gao-callan-2021-condenser}, RetroMAE~\citep{xiao-etal-2022-retromae}, and SimLM~\citep{wang-etal-2023-simlm} have enhanced the dense representation of BERT via customized architectures during language modeling.
Other works, including Contriever~\citep{izacard2022unsup}, GTE~\citep{li2023generaltextembeddingsmultistage}, E5~\citep{wang2024text} have further adapted two-stage contrastive learning.
These models are first trained with unsupervised or weakly supervised large-scale contrastive learning, followed by supervised contrastive learning with available relevance-judged data~\citep{nussbaum2024nomic, yu2024arcticembed}.
CDE~\citep{morris2024cde} further proposes a two-stage model architecture that integrates corpus-level information into document embeddings.

\subsection{LLM for Text Ranking}
On the other hand, recent large language models have shown strong potential in relevance modeling for text ranking.
Fine-tuning LLM as dense retriever models have shown significantly stronger effectiveness across various tasks and languages compared to smaller ones~\citep{wang-etal-2024-improving-text, muennighoff2024generative, springer2024repetition, li2024making}.
For example, RepLlama~\citep{ma2024repllama}, which uses straightforward supervised fine-tuning based on the Llama2-7B model, outperforms previous smaller retriever models that were based on multi-stage continuous pre-training, with a lower training cost.
This demonstrates the data efficiency and naturally strong generalization of LLM-based retrievers~\citep{luo-etal-2024-large}.

Moreover, instruction-following LLMs have also shown better performance when directly prompted as rerankers~\citep{ma2023zeroshot, sun-etal-2023-chatgpt}.
Reflecting the excel relevance understanding of large language models for retrieval.
In this work, we aim to leverage the characteristics of LLM-based ranking methods that are data-efficient and generalizable, shifting their high inference time costs into training time costs as data augmentation.

\subsection{Data Augmentation for Retriever}
InPars~\citep{bonifacio2022inpars} and Promptagator~\citep{promptagator} generate synthetic queries that align with given documents sampled from the task corpus, creating training data for retrieval corpora with limited human queries and judgments.
DRAGON~\citep{lin-etal-2023-train} enhances the robustness of dense retrievers by employing sentence cropping as pseudo-queries and generating augmented data based on retrieval results from multiple retrievers (e.g., sparse, multi-vector models).

With the emergence of LLMs, Mistral-E5~\citep{wang-etal-2024-improving-text} directly prompts an LLM to generate synthetic query-positive-negative triplets, using them as augmentation data to train a 7B LLM retriever across diverse text embedding tasks.
Gecko~\citep{lee2024gecko} takes a different approach by leveraging real documents: it generates synthetic queries from sampled real documents, retrieves top candidate passages, and uses an LLM to rerank them in pointwise way.
While these methods introduce various strategies for data augmentation in retrievers, they have not been systematically compared within a single framework where LLMs and corpora are controlled for fair comparison.
We explore various types of LLM-based data augmentation and evaluate their individual and combined effectiveness.

\subsection{Multilingual Retriever}
Multilingual capabilities are crucial for effective retrieval systems.
While numerous multilingual retrievers have been developed~\citep{izacard2022unsup, wang2024me5, zhang2024mgte, chen-etal-2024-m3}, they often face a trade-off between achieving strong performance in multilingual retrieval across various languages and preserves good English generalization performance on English retrieval.
While concurrent work ArcticEmbV2~\citep{yu2024arcticembed} also aims for effectiveness in both English and multilingual, they follow the previous training paradigm that firstly pretrain the model with contrastive learning over weakly supervised data pairs and then followed by supervised fine-tuning.
In our work, we address this challenge from a different view, by conducting data augmentation from LLM and using pruned LLM as the backbone of smaller retriever.

\section{Methods}

\subsection{Data Augmentation for Contrastive Dense Retriever Training}

Given a query $q$, a positive document $D^+$ relevant to the query, and a set of hard negative documents $\{D_{\text{HN}}\}$ that are similar to the positive document but are not highly relevant to the query, a dense retriever model is trained using the InfoNCE loss~\citep{oord2019infonce} as follows:
\begin{equation}\label{eq:obj}
\begin{split}
    &\mathcal{L}(q, D^+, \{D_{\text{N}}\}) \nonumber = -\log p(D=D^+\mid q) \\
    & = -\log \frac{\exp(\text{Sim}(q, D^+)/\tau)}{\sum\limits_{D_i \in \{D^+ \} \cup \{D_\text{N}\}} \exp(\text{Sim}(q, D_i)/\tau)},
\end{split}
\end{equation}

\noindent where $\{D_\text{N}\}$ is the union of the hard negative documents $\{D_\text{HN}\}$ for each query and in-batch negative documents, which are positive or hard negatives from other queries in the same training batch. The similarity $\text{Sim}(Q, D)$ is commonly computed as the cosine similarity between the embedding vectors of the query and document.

Data augmentation for dense retrieval focuses on creating triplets of queries $q$, positive documents $D_\text{P}$, and hard negative documents $\{D_\text{HN}\}$.
In this work, we make the following assumptions regarding available resources for data augmentation:

\begin{itemize}
    \item \textbf{Initial Supervised Data} ($D_{\text{sft}}$): A commonly accessible general-domain retrieval dataset.
    \item \textbf{Large Retrieval Model} ($\text{LLM}_{\text{Ret}}$): An LLM-based retrieval model, fine-tuned on $D_{\text{sft}}$.
    \item \textbf{Instruction-following LLM} ($\text{LLM}_{\text{Inst}}$): An LLM with strong instruction-following capability that can generate synthetic data reflecting its relevance preferences.
    \item \textbf{Large Corpus} ($C$): A diverse or multilingual document corpus that serves as the basis for synthetic query generation and relevance assessment.
\end{itemize}

\noindent With the above assumption, we explored various ways of utilizing LLM to conduct data augmentation for smaller retrievers, ranging from lower to higher computational costs for data creation.

\begin{figure*}
    \centering
    \includegraphics[width=\textwidth]{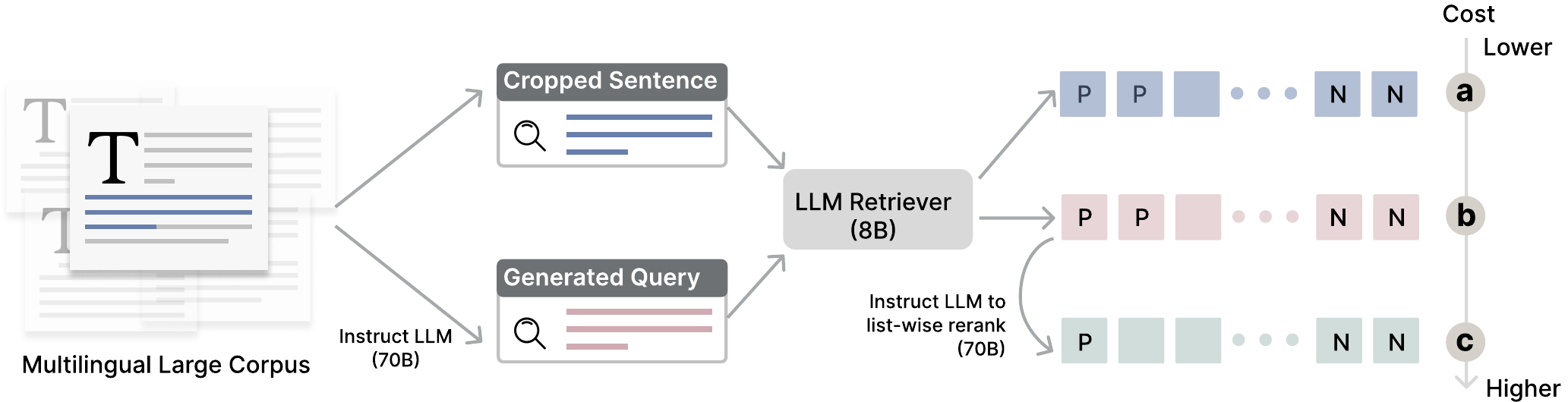}
    \caption{Methods to create data augmentation for smaller retriever with LLMs: (a) Using cropped sentences as queries, selecting the top-ranked documents from top-k retrieval as positives and the remaining as hard negatives. (b) Replacing cropped sentences with synthetic queries generated by prompting instruction-following LLM. (c) Refining retrieval results from the LLM retriever using an instruction-following LLM as a listwise reranker.}
    \label{fig:paradigm}
    \vspace{-0.4cm}
\end{figure*}

\subsubsection{Data Augmentation via \texorpdfstring{Llama-3.1$_\text{8B}$}{Llama-3.1-8B} Retriever}

Given an LLM-based retriever model, one of the simplest approaches to data augmentation, without relying on even larger LLMs, is to enable the smaller retriever to learn from the relevance preferences of the 8B embedding model $\text{LLM}_{\text{emb}}$.
Inspired by methods such as SPAR~\citep{chen-etal-2022-salient} and DRAGON~\citep{lin-etal-2023-train}, we begin with the corpus $C$.
For each document in $C$, we perform random sentence cropping to extract a smaller segment, which is treated as pseudo-query $q$.
These pseudo-queries, along with the full corpus, are encoded using the 8B retriever model.
Retrieval is then conducted for each pseudo-query $q$ to identify the top-$k$ candidate documents.
Among these candidates, the top $[1, m]$ documents are regarded as positive $D^+$, while the top $[k-n, k]$ documents are designated as hard negatives $D_{\text{HN}}$.
The process is illustrated in Figure~\ref{fig:paradigm}.a.
In this work, we set $k=50, m=10, n=20$.
This process generates an augmented dataset of size $|C|$, as each document contributes to the creation of pseudo-queries and their associated retrieval results.
The computational cost of this method is limited to encoding the $|C|$
 documents and pseudo-queries, followed by the retrieval of top-$k$
 candidates for each query.

\subsubsection{Synthetic Queries from \texorpdfstring{Llama-3.3$_\text{70B}$-Instruct}{Llama-3.3-70B-Instruct}}

The availability of instruction-following LLMs, such as Llama-3.3$_\text{70B}$-Instruct, enables the generation of synthetic queries that are more similar to real queries compared to those from random sentence cropping.
For each document in the corpus $C$, we prompt the LLM to generate a synthetic query $q$.
Similar to the above process, these LLM-generated queries are fed into the 8B $\text{LLM}_{\text{Ret}}$ to perform retrieval.
Based on the retrieval results, we can identify positive documents and hard negative documents for the synthetic queries as illustrated in Figure~\ref{fig:paradigm}.b.
While this approach retains the costs of encoding and retrieval, it introduces an additional computational cost associated with generating synthetic queries using the instruction-following LLM.

\subsubsection{LLM Ranking Preference from \texorpdfstring{Llama-3.3$_\text{70B}$-Instruct}{Llama-3.3-70B-Instruct}}
Instead of relying solely on the relevance preferences of the 8B embedding model, which are influenced by its fine-tuning on supervised data $D_{\text{sft}}$, the instruction-following LLM such as $\text{Llama-3.3}_{\text{70B}}$-Instruct can be further leveraged to refine relevance judgments.
Specifically, we prompt the LLM to perform listwise reranking of the top-$k$ candidates retrieved for each synthetic query, as illustrated in Figure~\ref{fig:paradigm}.c.
In this process, the LLM provides its relevance judgments by reranking the candidates.
The top-1 candidate after reranking is treated as the positive document $D^+$, while the top $[k-n, k]$ candidates from the reranked list are designated as hard negatives $D_{\text{HN}}$.
In our experiments, we set $k=20, n=10$.
This listwise reranking approach aligns more closely with how humans select the most relevant one among multiple candidates.
This data augmentation incurs additional computational costs of prompting the instruction-following LLM for reranking $k$ candidates for every query.

In practice, having the data augmentation from LLM listwise rerank can further improve the LLM$_\text{Ret}$ by combining the augmented data with the initial supervised data $D_\text{sft}$.
We sampled LLM listwise rerank augmented data as the same amount of $D_\text{sft}$ to re-train the LLM$_\text{Ret}$.
The effectiveness of this operation is further analyzed in Section.~\ref{ablation:data}.

\subsubsection{Triplet Generation from \texorpdfstring{Llama-3.3$_\text{70B}$-Instruct}{Llama-3.3-70B-Instruct}}

Another approach to leverage the LLM's relevance preferences for data augmentation is to directly prompt the LLM to generate triplets consisting of a query, a positive document, and a hard negative document.
This approach does not rely on a pre-existing corpus to provide seed documents.
Following Mistral-E5~\citep{wang-etal-2024-improving-text}, but adhering to our controlled data augmentation framework (i.e., creating the same amount of augmentation data with the same LLM), we first prompt the LLM to brainstorm $|C|$ retrieval tasks.
Each task includes a retrieval scenario $t$, a query $q$, and its context.
Based on the task and query, the LLM is then prompted to generate a corresponding positive document and a hard negative document.
While this method appears promising in theory, our experiments revealed that purely synthetic triplet data generated in this manner does not substantially improve the training of smaller retriever models.
Detailed analyses can be found in Section~\ref{ablation:data}.


\subsection{Pruning}
Previous pre-LLM-era retriever models predominantly utilized encoder-only architectures, such as BERT-base for English retrieval and XLM-RoBERTa-Large for multilingual retrieval.
In this work, in addition to leveraging LLMs for data augmentation, we investigate whether recent decoder-only LLMs can provide better backbones for smaller retriever models.
We perform structured pruning on an LLM to obtain models with non-embedding parameter sizes of 0.1B and 0.3B, making them comparable to BERT-base and XLM-RoBERTa-Large, respectively.
Specifically, we initialize the pruning process with Llama3.2$_\text{1B}$, itself a pruned version of Llama3.1$_\text{8B}$.
Following the methodology from ShearedLlama~\citep{xia2024sheared}, the pruning process is performed in two stages.

In the pruning stage, a parameter mask is learned to selectively prune the model, formulated as a constrained optimization problem.
Pruning masks $z$ are applied to model hard concrete distributions~\citep{louizos2018learningsparseneuralnetworks}.
Constraints are enforced via Lagrange multipliers to ensure the resulting model adheres to the target architecture.
For example, the loss function applied to a layer for a target number of attention heads $H_{\mathcal{T}}$ is defined as~\cite{xia2024sheared}:
\begin{align*}
\tilde{\mathcal{L}}^{\text{head}}(\lambda, \phi, z) &= 
\lambda^{\text{head}} \cdot ( \sum_{i} z^{\text{head}}_i - H_{\mathcal{T}} ) \notag \\
& + \phi^{\text{head}} \cdot ( \sum_{i} z^{\text{head}}_i - H_{\mathcal{T}} )^2
\end{align*}

\noindent Similarly, the full pruning loss integrates constraints on other structural components, including the layer mask $z^{\text{layer}}$, hidden dimension mask $z^{\text{hidden}}$, attention head mask $z^{\text{head}}$, and intermediate dimension mask $z^{\text{int}}$. These are combined with the standard language modeling objective as:
\begin{align*}
& \mathcal{L}_{\text{prune}}(\theta, z, \lambda, \phi) = 
\mathcal{L}(\theta, z) \\ & + 
\sum_{j=1}^{L_S} \tilde{\mathcal{L}}^{\text{head}}_j + 
\sum_{j=1}^{L_S} \tilde{\mathcal{L}}^{\text{int}}_j \notag + \tilde{\mathcal{L}}^{\text{layer}} + 
\tilde{\mathcal{L}}^{\text{hidden}}
\end{align*}

\noindent This is followed by a continuous pretraining stage, during which the pruned model is further trained only on the language modeling objective to recover its performance.

Pruning from an LLM offers several potential advantages compared to training traditional pre-trained language models.
First, it allows us to leverage the latest advancements in LLMs, which are trained on large-scale, high-quality datasets and exhibit strong generalization and multilingual capabilities.
Secondly, it supports longer contexts than earlier models, allowing for improved handling of retrieval scenarios requiring extended input sequences. 
Thirdly, the pruning process provides the flexibility to tailor model sizes based on specific deployment needs.

\section{Experiment Setup}
\subsection{Fine-tuning Data}
\label{ft-data}
Controlling the supervised fine-tuning data is critical for ensuring a fair comparison across methods when studying the generalizability of retrieval models.
BEIR~\citep{thakur2021beir} was originally designed for zero-shot evaluation, encouraging the use of MS MARCO Passage Retrieval as the sole fine-tuning dataset.
However, many recent retrievers incorporate supervised data from the evaluation tasks, making the evaluation not entirely zero-shot.
To balance fairness in assessing model generalization while maintaining adequate baselines for comparison, we follow the fine-tuning data setup of E5~\citep{wang2024text}.
This setup includes general-domain retrieval datasets
but not include fine-tuning data for domain-specific retrieval tasks such as financial QA or scientific document retrieval.
For our experiments, we use the open-source replication of the E5 fine-tuning data~\citep{li2024making}.

\subsection{Data Augmentation}
For the LLM retriever model $\text{LLM}_{\text{ret}}$, we initialize it with Llama3.1$_{\text{8B}}$ and first fine-tune it following the training recipe of RepLlama~\citep{ma2024repllama} for one epoch on the MS MARCO Passage Ranking training set~\citep{bajaj2018msmarcohumangenerated}.
We then further fine-tune it on the aforementioned E5 fine-tuning data to obtain an LLM retriever focusing on English retrieval.
We train another multilingual LLM retriever by continuous fine-tuning of the MS MARCO-trained LLM retriever using only the MIRACL~\citep{miracl} training data.
This allows us to better study generalization in the multilingual retrieval setting.

For the large corpus $C$ used in English data augmentation, we sample 25M documents from a diverse open web-crawled dataset.
For multilingual augmentation, we use a combination of multilingual Wikipedia and a multilingual web-crawled corpus covering 19 non-English languages, with each corpus containing 25M documents.
In both cases, we segment documents into text chunks of up to 256 tokens.


\subsection{Pruning}
We prune Llama3.2$_\text{1B}$ into 0.1B and 0.3B models, where the first pruning stage cost about 0.5 billion tokens and the second continuous pertaining stage cost 26 billion tokens. The data covering English and 19 non-English languages from web-crawled corpora.
The pruned models support a maximum context length of 8,192 tokens.
Detailed model configurations can be found in Appendix~\ref{app:prune}.

\begin{table*}[t]
\centering
\resizebox{\textwidth}{!}{
\begin{tabular}{lccccc|ccccc}
\toprule
& & & & & & \textbf{English} & \multicolumn{4}{c}{\textbf{Multilingual}} \\
\cmidrule(lr){7-7} \cmidrule(lr){8-11}
\textbf{Method} & \begin{tabular}[c]{@{}c@{}}\textbf{Non-Emb.} \\ \textbf{Param.}\end{tabular} & \begin{tabular}[c]{@{}c@{}}\textbf{Repre.} \\ \textbf{Dim.}\end{tabular} & \begin{tabular}[c]{@{}c@{}}\textbf{Contra.} \\ \textbf{Pretrain.}\end{tabular} & \begin{tabular}[c]{@{}c@{}}\textbf{Data} \\ \textbf{Aug.}\end{tabular} & \begin{tabular}[c]{@{}c@{}}\textbf{Multi.} \\ \textbf{Lang.}\end{tabular} & \textbf{BEIR (13)} & \textbf{MIRACL (18)} & \textbf{MTEB-FR (5)} & \textbf{MTEB-ZH (8)} & \textbf{MTEB-DE (4)} \\
\midrule
BM25               & -    & - & $\times$ & $\times$ & \checkmark       & 43.7 & 38.5 & - & - & - \\
\midrule
Contriever         & 86M  & 768 & \checkmark & $\times$ & $\times$       & 47.5 & -  & - & - & - \\
DRAGON             & 86M  & 768 &  $\times$   & $\checkmark$ & $\times$   & 50.2 & -  & - & - & - \\
E5-v2-base         & 86M  & 768 &  \checkmark & $\times$ & $\times$       & 51.9 & -  & - & - & - \\
bge-base-en-v1.5   & 86M  & 768 &  \checkmark & $\times$ & $\times$       & 55.0 & -  & - & - & - \\
mE5-base           & 86M  & 768 &  \checkmark & $\times$ & \checkmark     & 50.2 & 60.1 & 45.4 & 61.6 & 49.2 \\
mGTE-Dense         & 113M & 768 &  \checkmark & $\times$ & \checkmark     & 54.3 & 62.1 & 50.6 & \textbf{72.0} & 49.1 \\
ArcticEmb-v2-M     & 113M & 768 &  \checkmark & $\times$ & \checkmark     & \underline{56.9} & 59.2 & \underline{53.7} & 55.7 & 55.0 \\
\textbf{\ourmodel{}$_\text{0.1B}$}& 113M & 768 &  $\times$   & $\checkmark$ & \checkmark & \underline{56.9} & \underline{70.4} & 52.1 & 61.7 & \underline{55.1} \\
\midrule
E5-large-v2        & 303M & 1024 &  \checkmark & $\times$ & $\times$       & 52.1 & - & - & - & - \\
bge-large-en-v1.5  & 303M & 1024 & \checkmark & $\times$ & $\times$       & 56.1 & - & - & - & - \\
mE5-large          & 303M & 1024 & \checkmark & $\times$ & \checkmark     & 52.9 & 65.4 & 47.7 & 63.7 & 50.4 \\
mE5-Inst           & 303M & 1024 & \checkmark & $\checkmark$ & \checkmark & 54.1 & 66.0 & 49.9 & 64.2 & 52.5 \\
M3-BGE-Dense       & 303M & 1024 & \checkmark & $\times$ & \checkmark     & 50.0 & 69.2 & 48.6 & \underline{65.6} & 50.4 \\
ArcticEmb-v2-L     & 303M & 1024 & \checkmark & $\times$ & \checkmark     & 57.2 & 64.9 & 54.5 & 63.6 & \textbf{55.9} \\
\textbf{\ourmodel{}$_\text{0.3B}$} & 265M & 1024 & $\times$   & $\checkmark$ & \checkmark & \underline{58.0} & \underline{71.4} & \underline{54.8} & 63.0 & 55.6 \\
\midrule
Gecko              & 1B   & 768  & \checkmark & $\checkmark$ & \checkmark   & 58.0 & 56.2 & - & - & - \\
\textbf{\ourmodel{}$_\text{1B}$}  & 1B   & 2048 & $\times$     & $\checkmark$ & \checkmark & \textbf{59.1} & \textbf{71.7} & \textbf{57.6} & 63.7 & 56.2 \\
\textbf{\ourmodel{}$_\text{1B}$} (768d)  & 1B   & 768 & $\times$     & $\checkmark$ & \checkmark & 58.5 & 70.9 & 56.5 & 62.8 & 55.8 \\
\midrule
MistralE5          & 7B   & 4096 & $\times$   & $\checkmark$ & \checkmark   & 59.0 & 62.2 &  -   &  - & - \\
\bottomrule
\end{tabular}
}
\caption{Effectiveness of \ourmodel{} compared to baseline methods (measured in nDCG@10).
For each method, we indicate the number of non-embedding parameters, the text embedding dimensionality, whether contrastive pretraining is needed, whether data augmentation is applied during supervised fine-tuning, and whether the retriever supports multilingual retrieval.
The notation ($x$) after a dataset name indicates the average value across $x$ subsets within the dataset.
Detailed results for each subset are provided in the Appendix ~\ref{detailed-results}.
We highlight the highest score for each dataset in bold and the highest score within each parameter level with an underscore. The notation (768d) indicates that we use the first 768 dimensions of representations from \ourmodel{}$_\text{1B}$, as our model is trained with MRL.
}
\label{tab:main}
\vspace{-0.3cm}
\end{table*}

\subsection{Training}
The full training data for the smaller retriever models consists of: (1) LLM augmented data based on cropped sentences. (2) 25M LLM retriever augmented data based on generated queries. (3) 25M Inst-LLM listwise reranker augmented data based on generated queries.
These three types of data augmentation are applied to all sources, including English web-crawl corpora, multilingual Wikipedia, and multilingual web-crawl corpora (denoted as enWeb, mWiki, and mWeb respectively).
The sampling ratio of augmented data across these three sources is 2:1:1.
Additionally, the augmented data is mixed with the E5 supervised fine-tuning data, which contains approximately 2M instances.
See Appendix~\ref{sec:multilingual-ablation} for a detailed ablation study on different sampling ratios, and Appendix~\ref{app:train} for additional training details.

We train the model with each query paired with one positive document and seven hard negative documents for the 0.1B and 0.3B models and three hard negative documents for the 1B model.
We adopt the Matryoshka Representation Learning (MRL) during training to enable flexible dimensionality choice~\citep{kusupati2022matryoshka}.
See Appendix~\ref{sec:mrl} for the effectiveness of \ourmodel{} with different dimensionality configuration.

\subsection{Evaluation}
Our main evaluations are conducted on BEIR~\citep{thakur2021beir} and MIRACL~\citep{miracl}, to assess the generalization of dense retrievers and multilingual retrieval capability.
To further analyze the generalization of multilingual retrievers, we also evaluate on retrieval subsets of MTEB-FR~\citep{ciancone2024mtebfr}, MTEB-ZH~\citep{xiao2024cpack} and MTEB-DE.
To assess the effectiveness of long-context retrieval, which benefits from pruning an LLM, we evaluate on MLDR~\citep{chen-etal-2024-m3}, a benchmark for long-context multilingual retrieval across 13 languages. We also include evaluations on synthetic long-context retrieval benchmarks, NeedleRetrieval and PasskeyRetrieval, from the LongEmbed study~\cite{zhu-etal-2024-longembed}.
We use nDCG@10 as the metrics for all evaluations.

\subsection{Baselines}
We select representative baselines with similar retrieval task training data settings, as described in Sec.~\ref{ft-data}.
The major baselines include Contriever~\citep{izacard2022unsup}, DRAGON~\citep{lin-etal-2023-train}, E5~\citep{wang2024text}, BGE~\citep{xiao2024cpack}, mE5~\citep{wang2024me5}, BGE-M3~\citep{chen-etal-2024-m3}, mGTE~\citep{zhang2024mgte}, ArcticEmbV2~\citep{yu2024arcticembed}, Gecko~\citep{lee2024gecko}, and MistralE5~\citep{wang-etal-2024-improving-text}.

\section{Results}
\subsection{Generalization of Smaller Retrievers}

Table~\ref{tab:main} shows the performance of our \ourmodel{} variants on both English and multilingual retrieval tasks.
The results indicate that \ourmodel{} is a strong and generalizable retriever at different model sizes.
For example, \ourmodel{}$_\text{0.1B}$ achieves an nDCG@10 of 56.9 on BEIR, on par with ArcticEmb-v2-M, and outperforms other English-only and multilingual retrievers.
When scaling up to \ourmodel{}$_\text{0.3B}$, the score increases to 58.0, outperforming ArcticEmb-v2-L by 0.8 points and matching Gecko, which is a much larger 1B-parameter model.
Beyond English retrieval, \ourmodel{} exhibits strong multilingual capabilities.
On MIRACL, all \ourmodel{} variants (from 0.1B to 1B) outperform previous best models like M3-BGE-Dense, while also maintaining strong English retrieval performance.
This suggests that \ourmodel{} works well across different languages without losing effectiveness in English.

As discussed by~\citealp{lin-etal-2023-train}, there is often a trade-off between in-domain retrieval performance and generalization capability.
\ourmodel{} achieves very high in-domain multilingual effectiveness: for example, \ourmodel{}$_\text{0.3B}$ is 5.5 points higher than ArcticEmb-v2-L on MIRACL (which has training data included in $D_\text{sft}$).
However, it also maintains robust generalization performance in multilingual settings such as MTEB-FR.
On MTEB-ZH, \ourmodel{}$_\text{0.3B}$ performs slightly lower than ArcticEmb-v2, but the difference is within 1 point.
Overall, these results suggest \ourmodel{} is generalizable across retrieval tasks and languages.

\begin{table}[t]
\small
\centering
\resizebox{0.48\textwidth}{!}{
\begin{tabular}{lcccc|c}
\toprule
\multirow{2}{*}{\textbf{Method}} & \multirow{2}{*}{\textbf{Param.}} & \multirow{2}{*}{\textbf{L-CPT.}} & \multirow{2}{*}{\textbf{L-FT.}} & \multirow{2}{*}{\textbf{Max Len}} & \textbf{MLDR} \\
 &  &  &  &  & \textbf{Avg} \\
\midrule
BM25                     & - & $\times$ & $\times$ & $\infty$ & 53.6 \\
mE5-large                & 303M & $\times$ & $\times$ & 512 & 34.2 \\
M3-BGE-Dense             & 303M & \checkmark & $\times$ & 8192 & 45.0 \\
ArcticEmb-v2-M      & 113M & $\times$ & $\times$ & 8192 & 34.0 \\
\textbf{\ourmodel{}$_\text{0.1B}$}      & 113M & $\times$ & $\times$ & 8192 & 47.1 \\
\textbf{\ourmodel{}$_\text{0.3B}$}      & 265M & $\times$ & $\times$ & 8192 & 48.8 \\
\textbf{\ourmodel{}$_\text{1B}$}        & 1B   & $\times$ & $\times$ & 128k & \textbf{54.8} \\
\midrule
M3-BGE-Dense & 303M & \checkmark & \checkmark & 8192 & 52.5 \\
mGTE-Dense  & 113M & \checkmark & \checkmark & 8192 & 56.6  \\
\textbf{\ourmodel{}$_\text{0.1B}$-MLDR} & 113M & $\times$ & \checkmark & 8192 & 60.2 \\
\textbf{\ourmodel{}$_\text{0.3B}$-MLDR} & 265M & $\times$ & \checkmark & 8192 & 58.9 \\
\textbf{\ourmodel{}$_\text{1B}$-MLDR}   & 1B   & $\times$ & \checkmark & 128k & \textbf{62.3} \\
\bottomrule 
\end{tabular}
}
\caption{
Effectiveness of \ourmodel{} on the multilingual long-context retrieval task. 
L-CPT: Model has seen long-context data during contrastive pretraining. 
L-FT: Model has seen long-context data during supervised fine-tuning. 
Max Len: Maximum input length supported.
}
\label{tab:mldr}
\end{table}

\begin{table}[t]
\centering
\small
\resizebox{0.48\textwidth}{!}{
\begin{tabular}{lccccccc}
\toprule
\textbf{Model} & \textbf{Size} & \textbf{C-Max} & \textbf{LM-Max} & \textbf{512} & \textbf{4096} & \textbf{8192} & \textbf{32768} \\
\midrule
E5-v2-base          & 0.1B & 512  & 512   & 88.0 & 13.8 & 7.3  & 6.2 \\
E5-v2-base-RoPE     & 0.1B & 512  & 512   & 92.6 & 11.3 & 6.3  & 6.3 \\
\textbf{\ourmodel{}$_\text{0.1B}$}           & 0.1B & 256  & 8192  & 93.8 & 48.5 & 32.9 & 8.2 \\
BGE-M3-Dense        & 0.3B & 8192 & 8192  & 84.4 & 49.9 & 26.2 & 12.6 \\
\textbf{\ourmodel{}$_\text{0.3B}$}          & 0.3B & 256  & 8192  & 92.3 & 48.0 & 25.6 & 14.7 \\
\textbf{\ourmodel{}$_\text{1B}$}             & 1B   & 256  & 128k  & \textbf{95.2} & \textbf{77.8} & \textbf{53.8} & \textbf{19.0} \\
MistralE5          & 7B   & 512  & 4096  & 93.4 & 68.7 & 30.4 & 6.6 \\
\bottomrule
\end{tabular}}
\caption{Effectiveness of \ourmodel{} on synthetic long-context retrieval task NeedleRetrieval (nDCG@10) across different max input lengths. C-Max denotes the maximum input length during contrastive learning, and LM-Max denotes the maximum input length during language modeling.
Full results of NeedleRetrieval and PasskeyRetrieval are provided in Appendix~\ref{detailed-results}.}
\vspace{-0.4cm}
\label{tab:needle-retrieval}
\end{table}

\subsection{Pruned LLM as Retriever Backbone}

Pruning a state-of-the-art LLM to create smaller retriever backbones offers two key advantages.
First, it helps preserve multilingual capability.
Most existing retrievers at the 0.1B parameter scale use \texttt{bert-base-uncased} as their backbone.
While these models achieve strong performance in English retrieval, they do not support multilingual retrieval.
By pruning an LLM instead, we achieve strong English retrieval effectiveness while retaining its multilinguality with only a small amount of multilingual web data (less than 10B tokens).
Second, as recent LLMs are designed to handle long contexts, pruning an LLM as the retriever backbone allows better long-context retrieval capabilities.
Table~\ref{tab:mldr} shows that even though \ourmodel{}’s fine-tuning data does not include MLDR training data, and \ourmodel{} is not trained with text beyond 256 tokens, it still performs well in length extrapolation.
For example, \ourmodel{}$_\text{0.1B}$ achieves an nDCG@10 of 46.8 on MLDR, despite never being trained on long-context retrieval data.
Comparing \ourmodel{}$_\text{0.1B}$ to M3-BGE-Dense, which was trained with long-context data during contrastive pretraining but not fine-tuned on MLDR, \ourmodel{} outperforms it by 2.1 points.
This demonstrates the advantage of using a pruned LLM, which inherently supports longer contexts.

It is also important to note that BM25, a traditional lexical retrieval method, performs well in long-context retrieval.
However, after further fine-tuning \ourmodel{} on MLDR training data, it surpasses BM25 and other methods that have MLDR in training data.
This result shows the potential of further adapting \ourmodel{} to long-context multilingual retrieval tasks.

We also extend our evaluation to synthetic long-context retrieval tasks to further analyze effectiveness across different input lengths. Table~\ref{tab:needle-retrieval} presents results on NeedleRetrieval (PasskeyRetrieval follows a similar trend and is included in Appendix~\ref{detailed-results}). When comparing our pruned model to existing E5-v2-base and its variants that apply RoPE positional embeddings~\citep{zhu-etal-2024-longembed}, \ourmodel{}$_\text{0.1B}$ (with the same parameter size) demonstrates a clear advantage across all input lengths.
Compared to BGE-M3-Dense, which is trained with contrastive learning up to a maximum length of 8192, \ourmodel{}$\text{0.3B}$ outperforms it at input lengths of 512 and 32768.
At 4096 and 8192 tokens, \ourmodel{}$\text{0.3B}$ slightly lags behind, suggesting that training \ourmodel{} with even longer context lengths could further enhance its performance.
\ourmodel{}$\text{1B}$ achieves the highest scores across all input lengths.
We attribute this to its longer maximum input length inherited from its original language modeling, showing the potential of increasing context length during the pruning stage.

\begin{figure}[t]
    \centering
        \includegraphics[width=0.48\textwidth]{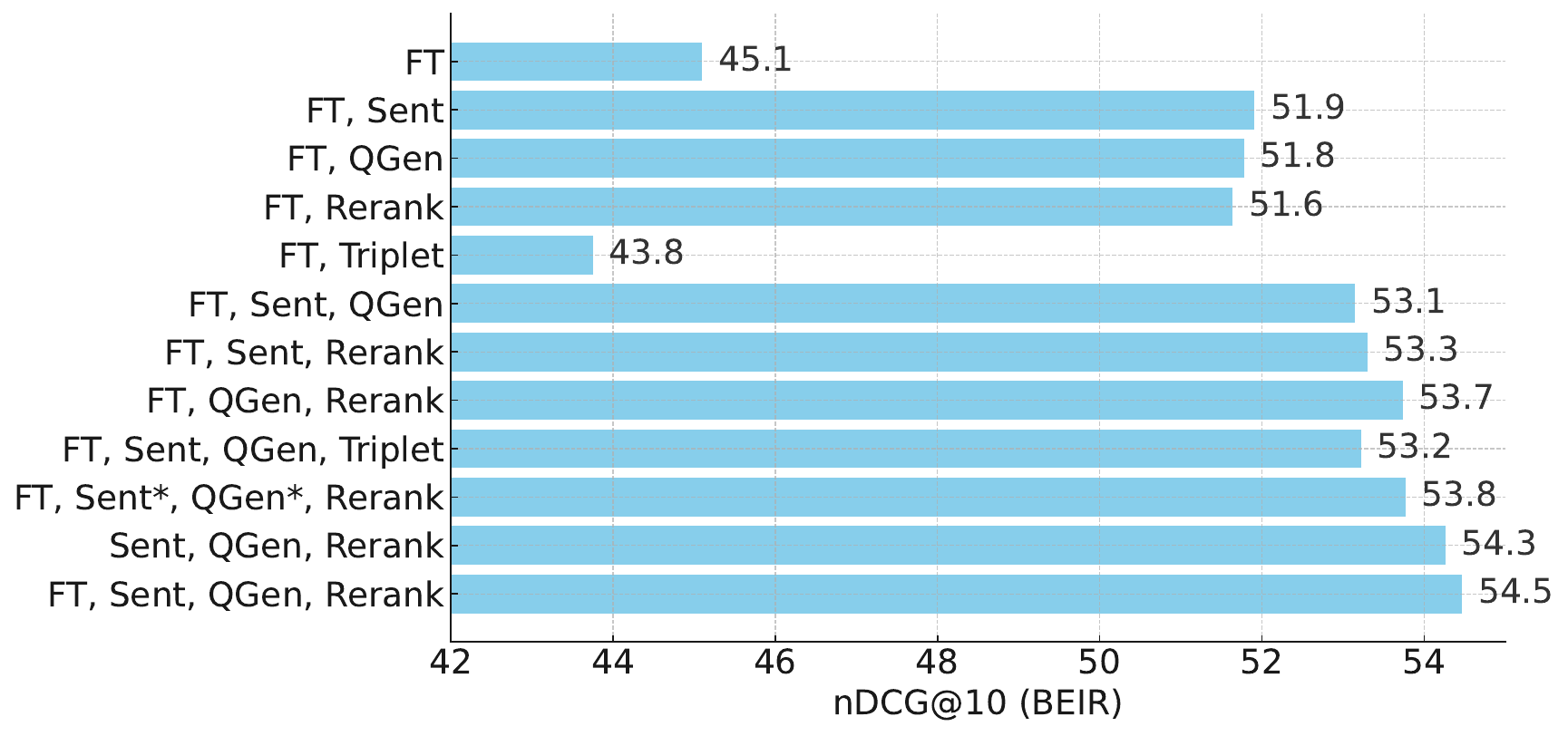}
        \caption{Effectiveness of different data augmentation combinations FT: SFT data, Sent: augmentation data based on cropped sentence query, QGen: augmentation data based on LLM generated query, Rerank: augmentation data based on LLM rerank. Triplet: augmentation data based on LLM triplet generation.). The model is trained based on 0.1B backbone, using only the English data augmentation and with 1 hard negative per query.}
        \vspace{-0.5cm}
    \label{fig:ablation-data}
\end{figure}

\section{Analysis and Ablation Study}

\subsection{Effectiveness of Data Augmentation}
\label{ablation:data}
Figure~\ref{fig:ablation-data} illustrates the effectiveness of different data augmentation combinations.
We observe that directly fine-tuning the model without data augmentation results in poor generalization performance. 
Incorporating any form of LLM-based data augmentation significantly improves BEIR performance, with one exception: directly prompting Llama3.3$_\text{70B}$-Instruct to generate synthetic triplets (queries, positive documents, and negative documents) does not yield meaningful improvements.
This suggests that training a smaller retriever model benefits more from using real-world documents.

Moreover, combining multiple types of data augmentation further enhances effectiveness beyond using any single augmentation method alone.
The highest performance is achieved when all three types of data augmentation are combined. 
Notably, when all augmentation strategies are applied together, the importance of fine-tuning data is diminishing, showing the effectiveness of our data augmentation approach.
The data point noted by [FT, Sent$^*$, QGen$^*$, Rerank] shows the performance of using LLM$_\text{Ret}$ without further improvement from LLM listwise rerank augmentation.
Its lower effectiveness compared to the final combination underscores that incorporating LLM-based rerank augmentation enhances the performance of LLM$_\text{Ret}$ and further improving the effectiveness of the smaller retriever model.

In practice, the computational cost estimation of creating Sent, QGen and Rerank data augmentation is in the order of 0.1k, 1k and 10k GPU hours respectively, where cost is dominated by the expensive LLM ranking for full data augmentation. 
On the other hand, we show that having Sent and QGen achieves a reasonable performance compared to using all 3 augmentations (53.1 vs. 54.5 on BEIR as shown in Figure~\ref{fig:ablation-data}). 
As a result, the computational cost of synthetic data generation can be lowered by an order of magnitude by omitting the LLM Ranking data, at the expense of losing around 2-3\% of retrieval effectiveness.

\subsection{Effectiveness of Model Backbone}

In Table~\ref{tab:backbone}, we compare the effectiveness of using a pruned Llama model as the retriever backbone against small encoder models.
At the 0.1B scale, the pruned model outperforms BERT by approximately 1 point on average.
Similarly, at the 0.3B scale, the pruned model surpasses XLM-RoBERTa-Large by about 1.5 points.
This demonstrates the effectiveness of using pruned-decoder-only LLM as a retriever backbone for text encoding tasks.
Additionally, the 0.1B pruned model performs slightly better than ModernBERT, a recently developed encoder-only model.
However, unlike ModernBERT, our approach retains multilingual support and leverages existing LLM pretraining, dropping the need to train the backbone from scratch.

\begin{table}[t]
    \centering
    \resizebox{0.41\textwidth}{!}{
    \begin{tabular}{l c c}
        \toprule
        \textbf{Backbone} & \textbf{Param.} & \textbf{BEIR} \\
        \midrule
        BERT & 0.1B & 53.50 \\
        ModernBERT & 0.1B & 54.22 \\
        Llama3.2$_{\text{1B}\rightarrow\text{0.1B}}$ & 0.1B & 54.47 \\
\midrule
        XLM-RoBERTa-Large & 0.3B & 54.74 \\
        Llama3.2$_{\text{1B}\rightarrow\text{0.3B}}$ & 0.3B & 56.14 \\
        \bottomrule
    \end{tabular}
    }
    \caption{Effectiveness of using pruned Llama3.2 as smaller retriever backbone compares to pre-LLM-era or recent encoder-only backbone. The models are trained using only the English data augmentation and with 1 hard negative per query.}
    \label{tab:backbone}
\end{table}

\begin{table}[t]
    \centering
    \resizebox{0.43\textwidth}{!}{
    \begin{tabular}{ccc c}
        \toprule
        \textbf{Model Size} & \textbf{Attention} & \textbf{Pooling} & \textbf{BEIR} \\
        \midrule
        0.1B & Bi-direction  & Mean  & 54.47 \\
             & Bi-direction  & EOS   & 54.37 \\
             & Uni-direction & Mean  & 53.88 \\
             & Uni-direction & EOS   & 53.58 \\
        \midrule
        0.3B & Bi-direction  & Mean  & 56.14 \\
             & Bi-direction  & EOS   & 55.85 \\
             & Uni-direction & Mean  & 55.18 \\
             & Uni-direction & EOS   & 54.79 \\
        \bottomrule
    \end{tabular}%
    }
    \caption{Impact of different attention and pooling mechanisms for the smaller retriever.
    The model is trained using only the English data augmentation and with 1 hard negative per query.}
    \label{tab:attention_ablation}
    \vspace{-0.4cm}
\end{table}

\subsection{Attention and Pooling Mechanism}
In Table~\ref{tab:attention_ablation}, we analyze how the attention mechanism and pooling strategy affect retrieval performance when training the pruned model as a text encoder.
It shows that bi-directional attention outperforms uni-directional attention.
While mean pooling yields higher scores than last-token pooling, the impact of the attention mechanism is greater than that of the pooling strategy.
Even with massive augmented training data, uni-directional attention remains a limiting factor.
However, simply enabling bi-directional attention allows the small decoder-only model to function more effectively.

\subsection{Cross-lingual Generalization}
\label{sec:cross-lingual}
In Table~\ref{tab:crosslingual_ablation}, we analyze how our model generalizes to zero-shot languages. 
The models are trained using English data augmentation and evaluated on languages that were not explicitly included in the fine-tuning stage.
First, we examine German (de), a higher-resource language.
The results show a clear trend where zero-shot effectiveness improves as the model size increases, suggesting that scaling up enhances cross-lingual generalization.
For Yoruba (yo), an interesting pattern emerges: the 0.3B pruned model outperforms the larger 1B model.
This may be due to the fact that the 1B model was not well-trained in Yoruba.
The pruning stage of our approach includes \texttt{yo} data, leading to stronger performance in this language.
In contrast, Polish (pl), which was not covered in either the fine-tuning or pruning stages, shows a noticeable performance gap compared to the 1B model. 
This shows the importance of including a language during pruning, as exposure at this stage significantly benefits zero-shot retrieval effectiveness. 

\begin{table}[t]
    \centering
\resizebox{0.48\textwidth}{!}{%
    \begin{tabular}{lccc}
        \toprule
        \textbf{Backbone} & \textbf{MIRACL-de} & \textbf{MIRACL-yo} & \textbf{MTEB-pl} \\
        \hline
        ${\text{1B}\rightarrow\text{0.1B}}$ & 45.48 & 68.77 & 32.38 \\
        ${\text{1B}\rightarrow\text{0.3B}}$ & 55.83 & 83.85 & 36.85 \\
        ${\text{1B}}$ & 58.20 & 76.20 & 51.08 \\
        \hline
    \end{tabular}
    }
    \caption{Cross-lingual generalization performance of models trained with English data augmentation, evaluated on zero-shot languages.
    DE and YO are seen during the pruning stage, while PL is unseen.
    For MTEB-pl, results are averaged over 11 retrieval tasks.}
    \vspace{-0.2cm}
    \label{tab:crosslingual_ablation}
\end{table}

\section{Conclusion}
We introduce \ourmodel{}, a training framework that leverages large language models to train smaller, generalizable dense retrievers by cohesively integrating LLM pruning and diverse LLM data augmentation.
\ourmodel{} achieves strong performance across English and multilingual retrieval tasks, enabling the training of smaller retrievers to improve together with advancements in LLMs.

\clearpage
\section*{Limitations}
While \ourmodel{} achieves strong retrieval effectiveness across English and multilingual tasks, several areas remain open for further investigation.

Firstly, the scope of language support.
As observed in Section~\ref{sec:cross-lingual}, including a language during the pruning stage is crucial for enabling the smaller model to generalize well to that language. 
While the 0.1B and 0.3B variants of \ourmodel{} covers 20 languages, expanding this coverage could improve performance for low-resource languages that lack sufficient contrastive learning data.
A more comprehensive pruning strategy, incorporating additional languages, would likely enhance zero-shot multilingual retrieval.

Another limitation lies in the amount of supervised fine-tuning data. To maintain a fair evaluation of generalization, we followed the E5 fine-tuning setup, which does not include domain-specific retrieval tasks such as financial and medical.
However, incorporating a broader range of supervised datasets could further improve retrieval performance across diverse domains.

Additionally, \ourmodel{} is trained with up to 256 context length.
Although it demonstrates strong extrapolation potential in long-context retrieval, it is worth more exploration on how to better integrate the long-context training data into the data augmentation efficiently.
One possible approach is to organize training batches based on context length~\citep{chen-etal-2024-m3}.

Besides, \ourmodel{} follows a single-stage training approach, where the model is directly fine-tuned from a pruned LLM.
While this simplifies the pipeline and produces strong generalization, it remains an open question whether combining with multi-stage pertaining~\cite{yu2024arcticembed} or recently proposed multi-stage distillation~\cite{zhang2025jasperstelladistillationsota} will help further improve the effectiveness of \ourmodel{}.

Finally, \ourmodel{} focused on retrieval tasks.
Many recent models additionally optimize for broader text embedding tasks such as clustering and classification as well as instruction following or reasoning-intensive retrieval~\cite{lee2024gecko, wang2024me5, su-etal-2023-one, asai-etal-2023-task, weller-etal-2025-followir, shao2025reasonirtrainingretrieversreasoning}.
We leave further integrate supervised fine-tuning data and LLM data augmentation for these tasks into \ourmodel{} training framework as future work.

\section*{Ethics Statement}
This work complies with the ACL Ethics Policy. We declare that there are no ethical issues in this paper, to the best of our knowledge.

\section*{Acknowledgments}
We sincerely thank Sheng-Chieh Lin, Rulin Shao, John X. Morris, Luyu Gao, and Minghan Li for the insightful discussions throughout this work. We also extend our appreciation to the anonymous reviewers for their invaluable suggestions.

\bibliography{custom}

\newpage

\appendix
\section{Appendix}

\begin{figure}[h]
    \centering
        \centering
        \includegraphics[width=0.4\textwidth]{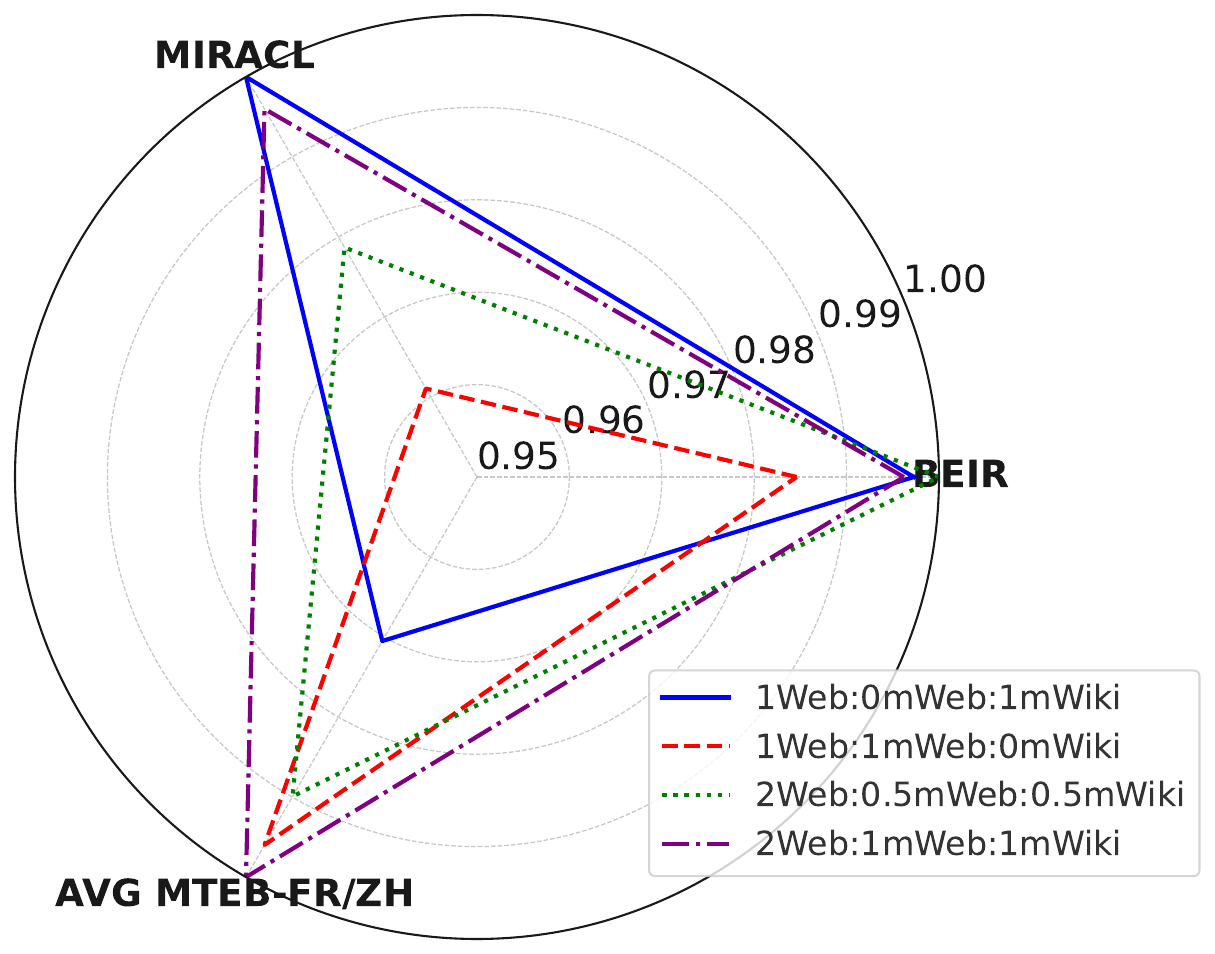}
        \caption{Effectiveness of different mixture ratios of English and multi-lingual data augmentation ratio for the data source of Web, mWeb and mWiki.
        The model is trained based on 0.1B backbone with 1 hard negative per query.}
        \label{fig:ablation-multilingual}
\end{figure}

\begin{figure}[t]
    \centering
        \includegraphics[width=0.43\textwidth]{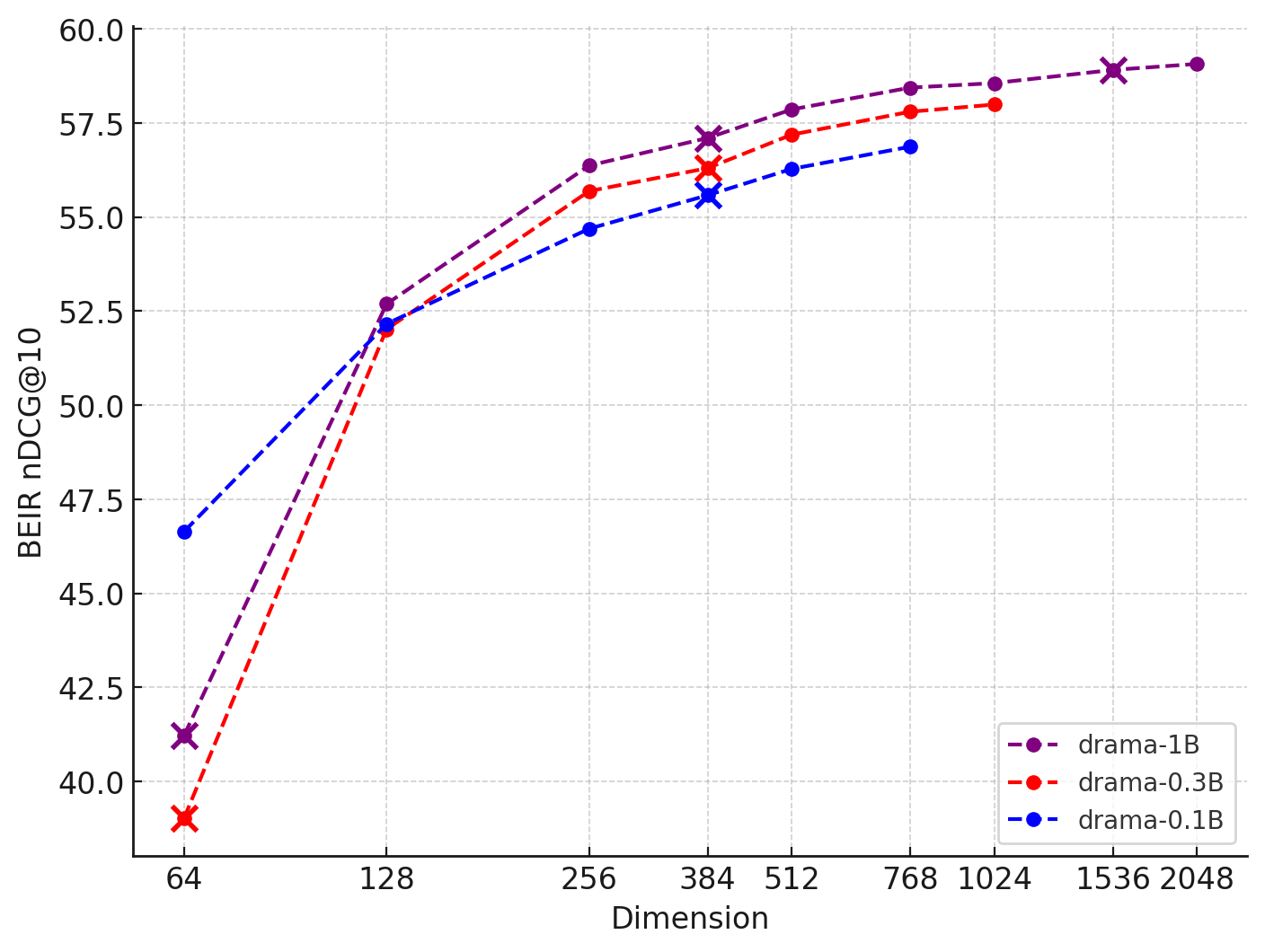}
        \caption{Effectiveness of \ourmodel{} across different text representation dimensions. Points marked with $\times$ indicate dimensionalities that were not explicitly optimized in the MRL process.}
        \label{fig:mrl}
\end{figure}

\subsection{Multilingual Data Balance}
\label{sec:multilingual-ablation}

Figure~\ref{fig:ablation-multilingual} illustrates how different mixtures of data sources affect effectiveness across English retrieval, in-domain multilingual retrieval, and multilingual generalization.
We observe that excluding mWeb negatively impacts multilingual generalization, likely due to overfitting on the Wikipedia corpus. Conversely, excluding mWiki leads to a drop in in-domain multilingual retrieval effectiveness. However, mixing both mWiki and mWeb enables strong performance across both in-domain effectiveness and multilingual generalization.
Additionally, we find that maintaining a 1:1 balance between English and multilingual data yields better overall performance than doubling the proportion of English data. While increasing the English proportion slightly improves BEIR effectiveness, it significantly weakens multilingual retrieval performance.
Overall, using a 1:1 ratio of English to multilingual data and incorporating augmentation data from both Wikipedia and web-crawled multilingual sources achieves the best trade-off, covering the largest area in the radar chart and ensuring robust performance across retrieval tasks.

\subsection{Matryoshka Representation Learning}
\label{sec:mrl}

In Figure~\ref{fig:mrl}, we compare the effectiveness of \ourmodel{} variants across different dense representation dimensionalities.
For dimensions larger than 256, the trend of model size scaling is clear---larger model achieves higher effectiveness. 
Additionally, text representations largely retain their effectiveness compared to using the full-dimensionality representation.
However, at 128 dimensions, the scaling trend is not guaranteed.
At 64 dimensions, the 0.1B model outperforms both the 0.3B and 1B models, likely because 64 dimensions were not a target setting during MRL training for the larger models.
In contrast, for dimensions 384 and 1536, despite also not being target dimensions for MRL, the effectiveness is well preserved.
This observation raises the importance of considering the range of target dimensionalities during MRL training to ensure effectiveness at test time.

\subsection{Pruned Model Configuration}
\label{app:prune}
Llama3.2-1B has a hidden size of 2048, an intermediate size of 8192, 32 attention heads, and 16 hidden layers. \textbf{\ourmodel{}$_\text{0.3B}$} reduces these dimensions to a hidden size of 1024, intermediate size of 4096, 16 attention heads, while maintaining 16 hidden layers. \textbf{\ourmodel{}$_\text{0.1B}$} further scales down to a hidden size of 768, intermediate size of 3072, 12 attention heads, and 12 hidden layers.

\subsection{Detailed Training Setup}
\label{app:train}

\paragraph{Model License:} Our LLM retriever is trained based on Llama3.1-8B follows Llama 3.1 Community License Agreement. Data augmentation based on Inst-LLM is based on Llama3.3-70B-Instruct follows Llama 3.3 Community License Agreement. Our backbone model is pruned based on Llama3.2-1B, following Llama 3.2 Community License Agreement.

\paragraph{Languages:} For pruning and data augmentation, our web crawl text corpora cover the following 20 languages:
English, Arabic, Bengali, Spanish, Persian, Finnish, French, Hindi, Indonesian, Japanese, Korean, Russian, Swahili, Telugu, Thai, Chinese, German, Yoruba, Italian, Portuguese.

\paragraph{Training:} The models are trained using the \texttt{dpr-scale}\footnote{\url{https://github.com/facebookresearch/dpr-scale}} codebase on 32 A100 GPUs over approximately two days. The training configurations for different model sizes are as follows:

\ourmodel{}$_\text{0.1B}$: Batch size of 2048, with each query paired with seven hard negatives.
\ourmodel{}$_\text{0.3B}$: Batch size of 1024, with each query paired with seven hard negatives.
\ourmodel{}$_\text{1B}$: Batch size of 256, with each query paired with three hard negatives.
All three variants are trained for 200,000 steps.

\subsection{Detailed Evaluation Results}
\label{detailed-results}
We use the \texttt{tevatron}\footnote{\url{https://github.com/texttron/tevatron}} codebase to evaluate BEIR and MIRACL.
For retrieval tasks in MTEB-FR/ZH/DE, we utilize the \texttt{mteb} codebase.
For BEIR and MIRACL, we set the maximum context length as 512 for both query and document following previous works.
For baselines, we adopt BEIR and MIRACL scores directly from the original works.
In MLDR, we reference baseline results from the mGTE work for mE5, BGE-M3, and mGTE. For Arctic-Embedding, we conduct the MLDR evaluation ourselves.
While some MTEB scores are reported in previous works, we observe version changes in certain datasets within MTEB-FR. To ensure consistency, we re-evaluate MTEB-FR/ZH/DE baselines ourselves.
We set the maximum context length as 1024 following~\cite{zhang2024mgte}.

The full evaluation results are presented in Table~\ref{tab:beir}, Table~\ref{tab:miracl}, Table~\ref{tab:mldr_full}, Table~\ref{tab:mteb_fr_retrieval}, Table~\ref{tab:mteb_zh_retrieval}, Table~\ref{tab:mteb_de_retrieval},
Table~\ref{tab:full-needleretrieval},
and Table~\ref{tab:full-passkeyretrieval}.

\begin{table*}[t]
\centering
\small
\resizebox{\textwidth}{!}{
\begin{tabular}{lccc|c|ccccccccccccc}
\toprule
\multirow{4}{*}{\textbf{Method}} & \multirow{4}{*}{\textbf{Param.}} & \multirow{4}{*}{\textbf{CPT}} & \multirow{4}{*}{\textbf{Multi.}} & \multicolumn{14}{c}{\textbf{BEIR} (nDCG@10)} \\
 &  &  &  & \textbf{Avg} & \begin{tabular}[c]{@{}c@{}}TREC-\\ COVID\end{tabular} & \begin{tabular}[c]{@{}c@{}}NF\\ Corpus \end{tabular} & \begin{tabular}[c]{@{}c@{}}Sci\\Fact\end{tabular} & \begin{tabular}[c]{@{}c@{}}SCI\\DOCS\end{tabular} & FiQA & \begin{tabular}[c]{@{}c@{}}Argu\\Ana\end{tabular} & \begin{tabular}[c]{@{}c@{}}Touche-\\ 2020\end{tabular} & \begin{tabular}[c]{@{}c@{}}DB\\Pedia\end{tabular} & \begin{tabular}[c]{@{}c@{}}Climate-\\ FEVER\end{tabular} & FEVER & NQ & \begin{tabular}[c]{@{}c@{}}Hotpot\\ QA\end{tabular} & Quora \\
\midrule
BM25               & -    & $\times$ & \checkmark & 43.7 & 59.5 & 32.2 & 67.9 & 14.9 & 23.6 & 39.7 & 44.2 & 31.8 & 16.5 & 65.1 & 30.5 & 63.3 & 78.9 \\
\midrule
Contriever         & 86M  & \checkmark & $\times$   & 47.5 & 59.6 & 32.8 & 67.7 & 16.5 & 32.9 & 44.6 & 23.0 & 41.3 & 23.7 & 75.8 & 49.8 & 63.8 & 86.5 \\
DRAGON             & 86M  & $\times$   & $\times$   & 50.2 & 75.9 & 33.9 & 67.9 & 15.9 & 35.6 & 46.9 & 26.3 & 41.7 & 22.7 & 78.1 & 53.7 & 66.2 & 87.5 \\
E5-v2-base         & 86M  & \checkmark & $\times$   & 51.9 & 69.6 & 35.4 & 71.9 & 18.7 & 39.9 & 44.5 & 26.4 & 42.2 & 26.6 & 85.0 & 58.2 & 69.2 & 86.6 \\
bge-base-en-v1.5   & 86M  & \checkmark & $\times$   & 55.0 & 78.1 & 37.4 & 74.0 & 21.7 & 40.6 & 63.6 & 25.7 & 40.8 & 31.2 & 86.3 & 54.1 & 72.6 & 88.9 \\
mE5-base           & 86M  & \checkmark & \checkmark & 50.2 & 69.7 & 32.5 & 69.3 & 17.2 & 38.2 & 44.2 & 21.4 & 40.4 & 23.9 & 79.4 & 60.0 & 68.6 & 87.6\\
mGTE-Dense               & 113M & \checkmark & \checkmark & 54.3 & 57.4 & 36.7 & 73.4 & 18.3 & 63.0 & 58.4 & 22.8 & 40.1 & 34.8 &  92.1 & 58.1 & 63.0 & 88.0 \\
ArcticEmb-v2-M     & 113M & \checkmark & \checkmark & 56.9 & 80.3 & 35.9 & 71.8 & 20.3 & 44.0 & 58.0 & 29.8 & 43.9 & 38.3 & 91.6 & 64.6 & 72.4 & 88.7 \\
\textbf{\ourmodel{}$_\text{0.1B}$}& 113M & $\times$   & \checkmark &  56.9 & 83.3 & 36.9 & 75.7 & 19.1 & 44.2 & 54.8 & 29.1 & 44.8 & 38.0 & 89.4 & 60.8 & 74.9 & 88.3 \\
\midrule
E5-large-v2        & 303M & \checkmark & $\times$   & 52.1 & 66.5 & 37.1 & 72.2 & 20.5 & 41.1 & 46.4 & 20.7 & 44.0 & 22.2 & 82.8 & 63.4 & 73.1 & 86.8 \\
bge-large-en-v1.5  & 303M & \checkmark & $\times$   & 56.1 & 74.8 & 38.1 & 74.6 & 22.6 & 45.0 & 63.5 & 24.8 & 44.1 & 36.6 & 87.2 & 55.0 & 74.1 & 89.1 \\
mE5-large          & 303M & \checkmark & \checkmark & 52.9 & 71.3 & 34.0 & 70.4 & 17.5 & 43.8 & 54.4 & 23.4 & 41.3 & 25.7 & 82.8 & 64.1 & 71.2 & 88.2 \\
mE5-Inst           & 303M & \checkmark & \checkmark & 54.1 & 82.0 & 35.5 & 71.9 & 18.7 & 47.7 & 58.4 & 27.2 & 38.4 & 29.9 & 78.0 & 57.8 & 69.3 & 89.1 \\
M3-BGE-Dense       & 303M & \checkmark & \checkmark & 50.0 & 55.6 & 31.4 & 64.4 & 16.4 & 41.3 & 54.0 & 22.6 & 39.8 & 24.2 & 81.4 & 60.6 & 69.4 & 88.6 \\
ArcticEmb-v2-L     & 303M & \checkmark & \checkmark & 57.2 & 83.9 & 35.3 & 70.6 & 20.2 & 45.5 & 59.2 & 29.5 & 43.4 & 43.5 & 91.9 & 63.7 & 68.2 & 89.0 \\
\textbf{\ourmodel{}$_\text{0.3B}$}& 265M & $\times$   & \checkmark & 58.0 & 83.8 & 37.9 & 76.1 & 19.7 & 46.9 & 54.1 & 28.1 & 47.7 & 41.9 & 89.5 & 64.1 & 75.6 & 88.4\\
\midrule
Gecko              & 1B   & \checkmark   & \checkmark & 58.0 & 82.6 & 40.3 & 75.4 & 20.4 & 59.2 & 62.2 & 25.9 & 47.1 & 33.2 & 87.0 & 61.3 & 71.3 & 88.2 \\
\textbf{\ourmodel{}$_\text{1B}$}  & 1B   & $\times$     & \checkmark & 59.1 & 85.8 & 37.6 & 77.9 & 20.7 & 50.6 & 53.5 & 29.6 & 50.0 & 38.7 & 89.9 & 67.3 & 77.4 & 88.7 \\
\textbf{\ourmodel{}$_\text{1B}$}(768d)  & 1B   & $\times$  & \checkmark & 58.4 & 85.2 & 37.1 & 77.5 &  20.7 & 50.2 & 53.1 & 29.0 & 49.2 & 37.9 & 89.5 & 66.5 & 75.5 & 88.5 \\

\midrule
MistralE5          & 7B   & $\times$   & \checkmark & 59.0 & 87.2 & 38.6 & 76.4 & 16.3 & 56.6 & 61.9 & 26.4 & 48.9 & 38.4 & 87.8 & 63.5 & 75.7 & 89.6 \\
\bottomrule 
\end{tabular}
}
\caption{Full BEIR evaluation of \ourmodel{}.}
\label{tab:beir}
\end{table*}

\begin{table*}[t]
\centering
\small
\resizebox{\textwidth}{!}{
\begin{tabular}{lc|c|cccccccccccccccccc}
\toprule
\multirow{3}{*}{\textbf{Method}} & \multirow{3}{*}{\textbf{Param.}} & \multicolumn{19}{c}{\textbf{MIRACL} (nDCG@10)} \\
 &  & \textbf{Avg} & ar & bn & en & es & fa & fi & fr & hi & id & ja & ko & ru & sw & te & th & zh & de & yo \\
\midrule
BM25 & - & 38.5 & 48.1 & 50.8 & 35.1 & 31.9 & 33.3 & 55.1 & 18.3 & 45.8 & 44.9 & 36.9 & 41.9 & 33.4 & 38.3 & 49.4 & 48.4 & 18.0 & 22.6 & 40.6 \\
\midrule
mE5-base    &           86M & 65.4 & 76.0 & 75.9 & 52.9 & 52.9 & 59.0 & 77.8 & 54.5 & 62.0 & 52.9 & 70.6 & 66.5 & 67.4 & 74.9 & 84.6 & 80.2 & 56.0 & 56.4 & 56.5 \\
mGTE-Dense   &  113M & 62.1 & 71.4 & 72.7 & 54.1 & 51.4 & 51.2 & 73.5 & 53.9 & 51.6 & 50.3 & 65.8 & 62.7 & 63.2 & 69.9 & 83.0 & 74.0 & 60.8 & 49.7 & 58.3 \\
ArcticEmb-v2-M & 113M & 59.2 &  -  &   -  &   -  &   -  &   -  &   -  &   -  &   -  &   -  &   -  &   -  &   -  &   -  &   -  &   -  &   -  &   -  &   -  \\
\textbf{\ourmodel{}$_\text{0.1B}$} & 113M & 70.4 & 80.5 & 74.5 & 56.3 & 61.4 & 62.8 & 78.9 & 62.2 & 61.9 & 58.0 & 74.2 & 70.5 & 72.3 & 77.1 & 81.5 & 80.4 & 64.8 & 62.3 & 88.5 \\
\midrule
mE5-large      & 303M & 60.1 & 71.6 & 70.2 & 51.2 & 51.5 & 57.4 & 74.4 & 49.7 & 58.4 & 51.1 & 64.7 & 62.2 & 61.5 & 71.1 & 75.2 & 75.2 & 51.5 & 43.4 & 42.3 \\
mE5-Inst  & 303M & 66.0 & 76.8 & 73.9 & 51.5 & 53.7 & 59.4 & 77.3 & 53.7 & 60.3 & 52.1 & 69.0 & 65.3 & 67.9 & 72.5 & 83.4 & 78.6 & 56.2 & 55.5 & 81.5 \\
M3-BGE-Dense         & 303M & 69.2 & 78.4 & 80.0 & 56.9 & 56.1 & 60.9 & 78.6 & 58.3 & 59.5 & 56.1 & 72.8 & 69.9 & 70.1 & 78.7 & 86.2 & 82.6 & 62.7 & 56.7 & 81.8 \\
ArcticEmb-v2-L & 303M & 64.9 &  -  &   -  &   -  &   -  &   -  &   -  &   -  &   -  &   -  &   -  &   -  &   -  &   -  &   -  &   -  &   -  &   -  &   -  \\
\textbf{\ourmodel{}$_\text{0.3B}$} & 265M  & 71.4 & 81.4 & 77.2 & 58.5 & 62.4 & 63.7 & 79.9 & 62.4 & 64.8 & 58.3 & 75.6 & 70.0 & 73.6 & 78.1 & 81.8 & 81.4 & 65.1 & 63.4 & 87.2 \\
\midrule
Gecko             & 1B & 56.2 &   -  &   -  &   -  &   -  &   -  &   -  &   -  &   -  &   -  &   -  &   -  &   -  &   -  &   -  &   -  &   -  &   -  &   -  \\
\textbf{\ourmodel{}$_\text{1B}$} & 1B & 71.7 & 81.1 & 76.6 & 58.4 & 62.2 & 64.5 & 80.9 & 62.8 & 65.7 & 58.7 & 76.4 & 69.3 & 74.6 & 77.6 & 80.6 & 81.8 & 68.2 & 63.9 & 88.1 \\
\midrule
MistralE5  & 7B & 62.2 & 73.3 & 70.3 & 57.3 & 52.2 & 52.1 & 74.7 & 55.2 & 52.1 & 52.7 & 66.8 & 61.8 & 67.7 & 68.4 & 73.9 & 74.0 & 54.0 & 54.0 & 58.8\\
\bottomrule 
\end{tabular}
}
\caption{Full MIRACL evaluation of \ourmodel{}.}
\label{tab:miracl}
\end{table*}

\begin{table*}[t]
\centering
\small
\resizebox{\textwidth}{!}{
\begin{tabular}{lcccc|c|ccccccccccccc}
\toprule
\multirow{3}{*}{\textbf{Method}} & \multirow{3}{*}{\textbf{Param.}} & \multirow{3}{*}{\textbf{L-CPT.}} & \multirow{3}{*}{\textbf{L-FT.}} & \multirow{3}{*}{\textbf{Max Len}} & \multicolumn{14}{c}{\textbf{MLDR} (nDCG@10)} \\
 &  &  &  &  & \textbf{Avg} & ar & de & en & es & fr & hi & it & ja & ko & pt & ru & th & zh \\
\midrule
BM25                     & - & $\times$ & $\times$ & $\infty$ & 53.6 & 45.1 & 52.6 & 57.0 & 78.0 & 75.7 & 43.7 & 70.9 & 36.2 & 25.7 & 82.6 & 61.3 & 33.6 & 34.6 \\
mE5-large                & 303M & $\times$ & $\times$ & 512 & 34.2 & 33.0 & 26.9 & 33.0 & 51.1 & 49.5 & 21.0 & 43.1 & 29.9 & 27.1 & 58.7 & 42.4 & 15.9 & 13.2 \\
M3-BGE-Dense             & 303M & \checkmark & $\times$ & 512 & 45.0 & 37.9 & 43.3 & 41.2 & 67.7 & 64.6 & 32.0 & 55.8 & 43.4 & 33.1 & 67.8 & 52.8 & 27.2 & 18.2 \\
ArcticEmb-v2-M           & 113M & $\times$ & $\times$ & 8192 & 34.0 & 15.9 & 35.4 & 32.4 & 67.0 & 63.9 & 15.2 & 56.8 & 10.0 & 17.7 & 66.1 & 42.9 & 11.2 & 7.4 \\
\textbf{\ourmodel{}$_\text{0.1B}$}      & 113M & $\times$ & $\times$ & 8192 & 47.1 & 39.6 & 46.7 & 40.6 & 73.4 & 72.9 & 27.0 & 57.9 & 44.5 & 36.2 & 69.5 & 55.3 & 30.0 & 19.0 \\
\textbf{\ourmodel{}$_\text{0.3B}$}      & 265M & $\times$ & $\times$ & 8192 & 48.8 & 42.9 & 49.8 & 44.1 & 75.1 & 73.2 & 30.1 & 62.3 & 42.4 & 34.4 & 71.2 & 58.4 & 32.7 & 17.7 \\
\textbf{\ourmodel{}$_\text{1B}$}        & 1B   & $\times$ & $\times$ & 128k & 54.8 & 49.0 & 53.2 & 51.1 & 79.2 & 76.9 & 36.7 & 68.4 & 53.3 & 43.3 & 77.5 & 63.0 & 37.1 & 24.0 \\
\midrule
M3-BGE-Dense & 303M & \checkmark & \checkmark & 8192 & 52.5 & 47.6 & 46.1 & 48.9 & 74.8 & 73.8 & 40.7 & 62.7 & 50.9 & 42.9 & 74.4 & 59.5 & 33.6 & 26.0 \\
mGTE-Dense  & 113M & \checkmark & \checkmark & 8192 & 56.6  & 55.0 & 54.9 & 51.0 & 81.2 & 76.2 & 45.2 & 66.7 & 52.1 & 46.7 & 79.1 & 64.2 & 35.3 & 27.4 \\
\textbf{\ourmodel{}$_\text{0.1B}$-MLDR} & 113M & $\times$ & \checkmark & 8192 & 60.2 & 60.6 & 55.3 & 56.6 & 84.0 & 81.3 & 43.6 & 72.2 & 55.9 & 48.7 & 82.3 & 73.8 & 38.8 & 29.1 \\
\textbf{\ourmodel{}$_\text{0.3B}$-MLDR} & 265M & $\times$ & \checkmark & 8192 & 58.9 & 58.2 & 53.1 & 57.0 & 83.1 & 81.0 & 39.9 & 71.0 & 54.9 & 47.5 & 80.8 & 71.8 & 39.2 & 28.7 \\
\textbf{\ourmodel{}$_\text{1B}$-MLDR}   & 1B   & $\times$ & \checkmark & 128k & 62.3 & 59.9 & 58.2 & 62.1 & 84.6 & 81.6 & 49.2 & 77.6 & 57.9 & 52.7 & 84.3 & 70.8 & 43.7 & 32.9 \\
\bottomrule 
\end{tabular}
}
\caption{
Full MLDR evaluation of \ourmodel{}.
}
\label{tab:mldr_full}
\end{table*}

\begin{table*}[t]
\centering
\small
\resizebox{0.9\textwidth}{!}{
\begin{tabular}{lc|c|ccccc}
\toprule
\multirow{3}{*}{\textbf{Method}} & \multirow{3}{*}{\textbf{Param.}} & \multicolumn{6}{c}{\textbf{MTEB-FR-Retrieval} (nDCG@10)} \\
 &  & \textbf{Avg} & \textbf{AlloprofRetrieval} & \textbf{BSARDRetrieval} & \textbf{MintakaRetrieval} & \textbf{SyntecRetrieval} & \textbf{XPQARetrieval} \\
\midrule
mE5-base    & 86M & 45.4 & 34.4 & 18.8 & 31.0 & 82.9 & 59.6 \\
mGTE-Dense   & 113M & 50.6 & 49.4 & 19.1 & 34.7 & 82.6 & 67.4 \\
ArcticEmb-v2-M & 113M & 53.7 & 54.6 & 18.4 & 31.4 & 89.8 & 74.4 \\
\textbf{\ourmodel{}$_\text{0.1B}$} & 113M & 52.1 & 51.9 & 24.7 & 26.7 & 85.5 & 71.5 \\
\midrule
mE5-large & 303M & 47.7 & 39.3 & 21.4 & 34.2 & 82.4 & 61.3 \\
mE5-Inst & 303M & 49.9 & 51.4 & 24.3 & 30.3 & 86.2 & 57.4 \\
M3-BGE-Dense & 303M & 48.6 & 48.3 & 16.6 & 22.9 & 84.5 & 70.9 \\
ArcticEmb-v2-L & 303M & 54.5 & 53.9 & 21.9 & 30.7 & 88.5 & 77.3 \\
\textbf{\ourmodel{}$_\text{0.3B}$} & 265M & 54.8 & 55.8 & 26.6 & 28.8 & 89.9 & 72.8 \\
\textbf{\ourmodel{}$_\text{1B}$} & 1B & 57.6 & 55.9 & 29.9 & 37.5 & 91.6 & 72.9 \\
\bottomrule 
\end{tabular}
}
\caption{Full MTEB-FR-Retrieval evaluation of \ourmodel{}.}
\label{tab:mteb_fr_retrieval}
\end{table*}

\begin{table*}[t]
\centering
\small
\resizebox{0.8\textwidth}{!}{
\begin{tabular}{lc|c|cccccccc}
\toprule
\multirow{3}{*}{\textbf{Method}} & \multirow{3}{*}{\textbf{Param.}} & \multicolumn{8}{c}{\textbf{MTEB-ZH-Retrieval} (nDCG@10)} \\
 &  & \textbf{Avg} & \textbf{Cmedqa} & \textbf{Covid} & \textbf{Du} & \textbf{Ecom} & \textbf{Medical} & \textbf{MMarco} & \textbf{T2} & \textbf{Video} \\
\midrule
mE5-base    & 86M & 61.6 & 27.2 & 73.5 & 81.7 & 54.2 & 48.4 & 76.0 & 70.8 & 61.3 \\
mGTE-Dense   & 113M & 72.0 & 43.8 & 81.0 & 87.5 & 64.8 & 61.9 & 79.4 & 84.7 & 72.8 \\
ArcticEmb-v2-M & 113M & 55.7 & 19.7 & 72.2 & 68.4 & 48.6 & 38.3 & 71.2 & 71.3 & 56.1 \\
\textbf{\ourmodel{}$_\text{0.1B}$} & 113M & 61.7 & 21.2 & 78.4 & 74.9 & 57.9 & 42.4 & 76.2 & 76.4 & 66.0 \\
\midrule
mE5-large      & 303M & 63.7 & 28.7 & 75.6 & 85.3 & 54.7 & 51.5 & 79.2 & 76.1 & 58.2 \\
mE5-Inst  & 303M & 64.2 & 33.9 & 76.1 & 85.2 & 53.7 & 56.2 & 78.6 & 82.9 & 47.2 \\
M3-BGE-Dense & 303M & 65.6 & 33.8 & 78.3 & 84.0 & 58.5 & 54.2 & 77.3 & 81.5 & 57.0 \\
ArcticEmb-v2-L & 303M & 63.6 & 27.8 & 78.8 & 78.4 & 56.4 & 51.1 & 78.4 & 79.7 & 58.6 \\
\textbf{\ourmodel{}$_\text{0.3B}$} & 265M & 63.0 & 21.2 & 78.4 & 74.9 & 57.9 & 42.4 & 76.2 & 76.4 & 66.0 \\
\textbf{\ourmodel{}$_\text{1B}$} & 1B & 63.7 & 23.6 & 76.1 & 77.8 & 60.1 & 45.8 & 79.4 & 79.0 & 67.8 \\
\bottomrule 
\end{tabular}
}
\caption{Full MTEB-ZH-Retrieval evaluation of \ourmodel{}.}
\label{tab:mteb_zh_retrieval}
\end{table*}

\begin{table*}[t]
\centering
\small
\resizebox{0.8\textwidth}{!}{
\begin{tabular}{lc|c|cccc}
\toprule
\multirow{3}{*}{\textbf{Method}} & \multirow{3}{*}{\textbf{Param.}} & \multicolumn{5}{c}{\textbf{MTEB-DE-Retrieval} (nDCG@10)} \\
 &  & \textbf{Avg} & \textbf{GerDaLIR} & \textbf{GermanDPR} & \textbf{GermanQuAD-Retrieval} & \textbf{XMarket} \\
\midrule
mE5-base    & 86M & 49.2 & 6.9 & 79.6 & 93.9 & 16.3 \\
mGTE-Dense   & 113M & 49.1 & 9.4 & 80.0 & 91.1 & 16.0 \\
ArcticEmb-v2-M & 113M & 55.0 & 16.1 & 81.8 & 94.4 & 27.6 \\
\textbf{\ourmodel{}$_\text{0.1B}$} & 113M & 55.1 & 15.4 & 82.8 & 95.9 & 26.2 \\
\midrule
mE5-large      & 303M & 50.4 & 6.5 & 82.9 & 94.6 & 17.5 \\
mE5-Inst  & 303M & 52.5 & 10.7 & 79.4 & 94.5 & 25.3 \\
M3-BGE-Dense & 303M & 50.4 & 10.9 & 82.5 & 95.1 & 13.1 \\
ArcticEmb-v2-L & 303M & 55.9 & 17.5 & 83.7 & 95.2 & 27.0 \\
\textbf{\ourmodel{}$_\text{0.3B}$} & 265M & 55.6 & 15.7 & 82.6 & 96.4 & 27.7 \\
\textbf{\ourmodel{}$_\text{1B}$} & 1B & 56.2 & 15.3 & 84.4 & 97.1 & 28.0 \\
\bottomrule 
\end{tabular}
}
\caption{Full MTEB-DE-Retrieval evaluation of \ourmodel{}.}
\label{tab:mteb_de_retrieval}
\end{table*}

\begin{table*}[t]
\centering
\small
\resizebox{0.9\textwidth}{!}{
\begin{tabular}{lcccccrrrrrr}
\toprule
\textbf{Model} & \textbf{Size} & \textbf{C-Max} & \textbf{LM-Max} & \textbf{256} & \textbf{512} & \textbf{1024} & \textbf{2048} & \textbf{4096} & \textbf{8192} & \textbf{16384} & \textbf{32768} \\
\midrule
E5-v2-base        & 0.1B & 512  & 512   & 91.8 & 88.0 & 40.8 & 21.4 & 13.8 & 7.3  & 2.7  & 6.2 \\
E5-v2-base-RoPE   & 0.1B & 512  & 512   & 96.3 & 92.6 & 44.0 & 20.0 & 11.3 & 6.3  & 3.8  & 6.3 \\
\textbf{\ourmodel{}$_\text{0.1B}$}        & 0.1B & 256  & 8192  & 95.2 & 93.8 & 92.8 & 84.2 & 48.5 & 32.9 & 24.7 & 8.2 \\
BGE-M3-Dense      & 0.3B & 8192 & 8192  & 94.2 & 84.4 & 80.0 & 53.6 & 49.9 & 26.2 & 30.4 & 12.6 \\
\textbf{\ourmodel{}$_\text{0.3B}$}       & 0.3B & 256  & 8192  & 95.4 & 92.3 & 88.2 & 71.7 & 48.0 & 25.6 & 15.4 & 14.7 \\
\textbf{\ourmodel{}$_\text{1B}$}          & 1B   & 256  & 128k  & 96.6 & 95.2 & 88.8 & 88.8 & 77.8 & 53.8 & 24.4 & 19.0 \\
MistralE5        & 7B   & 512  & 4096  & 97.4 & 93.4 & 86.9 & 74.8 & 68.7 & 30.4 & 21.3 & 6.6 \\
\bottomrule
\end{tabular}
}
\caption{Full NeedleRetrieval evaluation of \ourmodel{}.}
\label{tab:full-needleretrieval}
\end{table*}

\begin{table*}[ht]
\centering
\small
\resizebox{0.9\textwidth}{!}{
\begin{tabular}{lcccccrrrrrr}
\toprule
\textbf{Model} & \textbf{Size} & \textbf{C-Max} & \textbf{LM-Max} & \textbf{256} & \textbf{512} & \textbf{1024} & \textbf{2048} & \textbf{4096} & \textbf{8192} & \textbf{16384} & \textbf{32768} \\
\midrule
E5-v2-base        & 0.1B & 512  & 512   & 100.0 & 98.3 & 62.0 & 22.0 & 14.0 & 6.0  & 6.5  & 6.4 \\
E5-v2-base-RoPE   & 0.1B & 512  & 512   & 100.0 & 100.0 & 62.0 & 22.0 & 14.0 & 6.0  & 4.7  & 6.6 \\
\textbf{\ourmodel{}$_\text{0.1B}$}        & 0.1B & 256  & 8192  & 100.0 & 100.0 & 100.0 & 100.0 & 65.5 & 34.4 & 14.6 & 12.6 \\
BGE-M3-Dense      & 0.3B & 8192 & 8192  & 100.0 & 100.0 & 99.3 & 85.9 & 43.8 & 39.8 & 17.9 & 17.4 \\
\textbf{\ourmodel{}$_\text{0.3B}$}       & 0.3B & 256  & 8192  & 100.0 & 100.0 & 100.0 & 98.8 & 51.1 & 30.2 & 17.8 & 7.9 \\
\textbf{\ourmodel{}$_\text{1B}$}          & 1B   & 512  & 128k  & 100.0 & 100.0 & 100.0 & 100.0 & 100.0 & 96.2 & 48.0 & 16.0 \\
MistralE5        & 7B   & 512  & 4096  & 98.3  & 97.8  & 96.3  & 99.3  & 100.0 & 50.0 & 32.0 & 8.0 \\
\bottomrule
\end{tabular}
}
\caption{Full PasskeyRetrieval evaluation of \ourmodel{}.}
\label{tab:full-passkeyretrieval}
\end{table*}

\end{document}